\documentclass[final]{cvpr}

\usepackage{times}
\usepackage{epsfig}
\usepackage{graphicx}
\usepackage{amsmath}
\usepackage{amssymb}
\usepackage[title]{appendix}
\usepackage{multirow}
\usepackage{adjustbox}
\usepackage{blindtext}
\usepackage[table,xcdraw]{xcolor}
\usepackage{float}
\usepackage{enumitem}
\usepackage[table,xcdraw]{xcolor}
\usepackage{placeins}
\usepackage{xcolor}
\usepackage{subfigure}
\usepackage{dcolumn}
\newcolumntype{d}[1]{D{.}{.}{#1}}

\makeatletter
\newcolumntype{B}[3]{>{\boldmath\DC@{#1}{#2}{#3}}c<{\DC@end}}
\makeatother

\setlist{nosep} 

\restylefloat{table}

\newcommand{\matr}[1]{\mathbf{#1}} 

\usepackage[pagebackref=true,breaklinks=true,colorlinks,bookmarks=false]{hyperref}

\def\cvprPaperID{10320} 
\def\confYear{CVPR 2021}
\begin{document}

\title{Incremental Few-Shot Instance Segmentation}

\author{Dan Andrei Ganea\\
Utrecht University\\
{\tt\small dan.andrei.ganea@gmail.com}
\and
Bas Boom\\
Cyclomedia Technology\\
{\tt\small bboom@cyclomedia.com}
\and
Ronald Poppe\\
Utrecht University\\
{\tt\small r.w.poppe@uu.nl}
}
\maketitle
\thispagestyle{empty}


\begin{abstract}
    Few-shot instance segmentation methods are promising when labeled training data for novel classes is scarce. However, current approaches do not facilitate flexible addition of novel classes. They also require that examples of each class are provided at train and test time, which is memory intensive. In this paper, we address these limitations by presenting the first incremental approach to few-shot instance segmentation: iMTFA. We learn discriminative embeddings for object instances that are merged into class representatives. Storing embedding vectors rather than images effectively solves the memory overhead problem. We match these class embeddings at the RoI-level using cosine similarity. This allows us to add new classes without the need for further training or access to previous training data. In a series of experiments, we consistently outperform the current state-of-the-art. Moreover, the reduced memory requirements allow us to evaluate, for the first time, few-shot instance segmentation performance on all classes in COCO jointly\footnote{Code available at: \url{https://github.com/danganea/iMTFA}}. 
\end{abstract}

\section{Introduction}

Convolutional neural networks (CNNs) have led to state-of-the-art results for image classification \cite{krizhevsky2012imagenet, simonyan2014very}, object detection \cite{ren2015faster} and instance segmentation \cite{he2017mask}. In general, performance increases with network depth and training set size. While we can usually rely on large annotated databases for more general classes, adding a class for which we have little training data available is challenging. For example, we typically have a modest number of labeled training images when adding new classes for state-specific street furniture for self-driving cars, or types of weapons for automated detection in social media videos. Especially for instance segmentation, obtaining pixel-level annotations is costly.

\begin{figure}[htb]
	\begin{center}
		\includegraphics[width=1.0\linewidth]{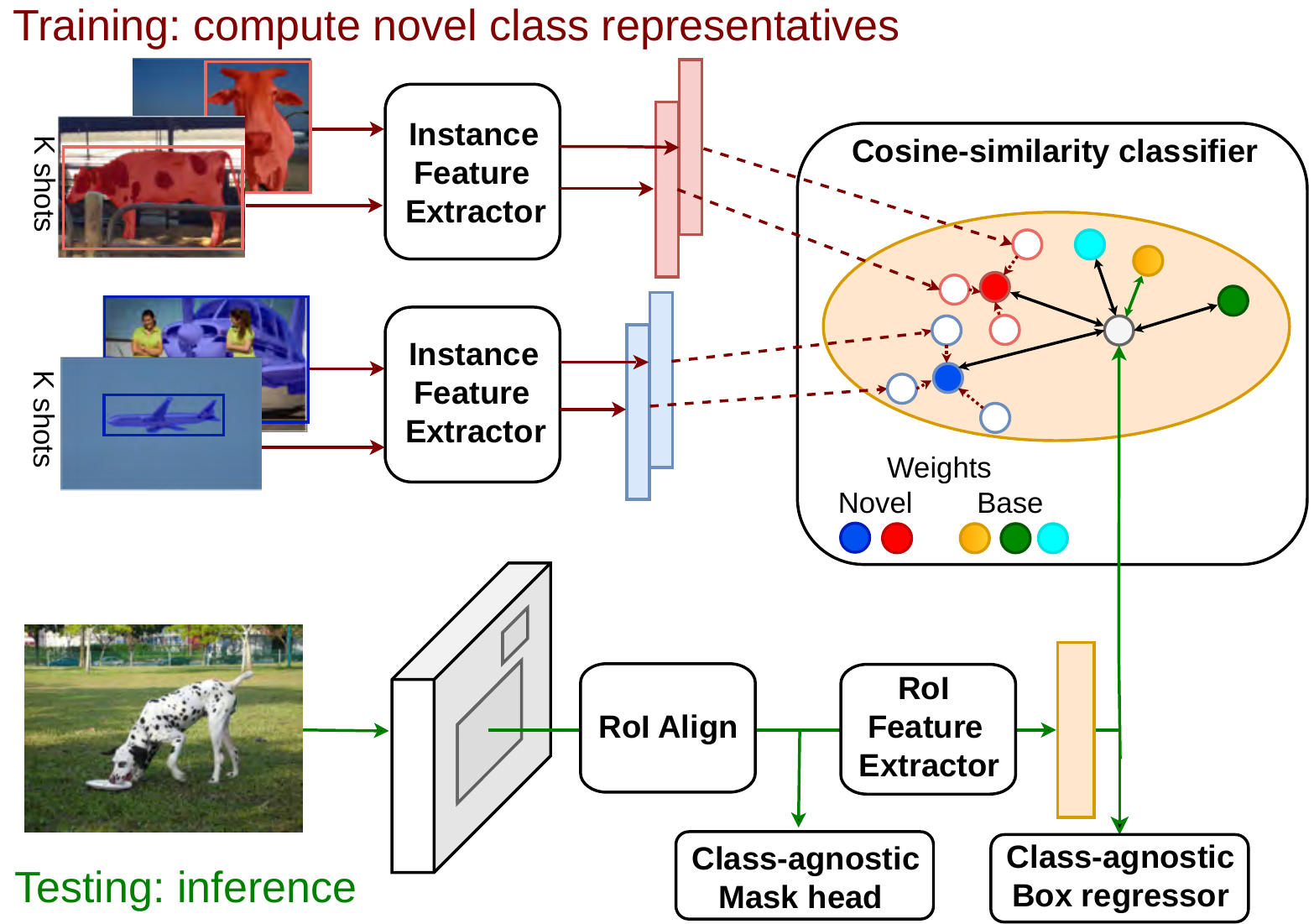}
	\end{center}
	\caption{\textbf{Incremental few-shot instance segmentation}. For all $K$ instances of each novel class, we produce vector embeddings using an Instance Feature Extractor. The average of these embeddings is stored as a per-class weight vector inside a cosine-similarity classifier. At test time (green), we compare the cosine distance embeddings of object proposal to the per-class weights.}
	\label{fig:firstimage}
\end{figure}

\textit{Few-shot learning} addresses the problem of learning with limited available data. Typically, one assumes the existence of a set of \textit{base classes}, for which there exist numerous training samples, and a disjoint set of \textit{novel classes}, for which training data is scarce ($K$ examples). The goal is to train a system to correctly classify $N$ classes: only the novel classes, or both novel and base classes jointly.

Compared to few-shot image classification, few-shot object detection (FSOD) and few-shot instance segmentation (FSIS) have received significantly less attention. While the few solutions that have been introduced show great promise, there is room for improvement in terms of practicality and accuracy. Often, long training procedures with both novel and base class samples are required \cite{kang2019few,wang2020frustratingly,yan2019meta}. This is unpractical when we flexibly want to add novel classes to a trained network. In \textit{incremental} few-shot learning, the addition of novel classes is independent from previous data, so computation time is reduced.

In this paper, we introduce the first incremental few-shot instance segmentation method: iMTFA (Figure~\ref{fig:firstimage}). We employ a two-stage training and fine-tuning approach based on Mask R-CNN~\cite{he2017mask}. The first stage trains the Mask R-CNN network. In the second stage, the fully-connected layers at the region of interest (RoI) level are re-purposed. Essentially, we transform a fixed feature extractor into an Instance Feature Extractor (IFE) that produces discriminative embeddings that are aligned with the per-class representatives. These embeddings are subsequently used as weights inside a cosine-similarity classifier.

Our approach has several advantages. First, it eliminates the need for extensive retraining procedures for new classes because these can be added incrementally. The IFE generates embeddings that are used as class representatives without requiring access to base classes. Because we predict localization and segmentation in a class-agnostic manner, these embeddings are all that is needed to add novel classes.

Second, in contrast with related methods  \cite{fan2020fgn,yan2019meta}, our mask predictor is class-agnostic. Similar to \cite{michaelis2018one}, no mask labels are needed for the addition of novel classes.

Third, our approach incurs no performance drawbacks at test time. We neither require additional memory for every class example \cite{fan2020fgn, yan2019meta} nor require these examples to be passed one-by-one (e.g., \cite{michaelis2018one}).

We make two main contributions:
\begin{itemize}
    \item We present the first incremental few-shot instance segmentation method: iMTFA. Our method outperforms the current state-of-the-art for FSIS as well as the current state-of-the-art in incremental FSOD.
    \item To compare between incremental and non-incremental methods, we extend an existing FSOD approach \cite{wang2020frustratingly} to the instance segmentation task (MTFA), and also demonstrate state-of-the-art results.
\end{itemize}

The remainder of the paper is structured as follows. We first discuss related work on few-shot learning and instance segmentation. We introduce our novel incremental and non-incremental methods in Section~\ref{sec:method}, and evaluate both extensively in Section~\ref{sec:experiments}. We conclude in Section~\ref{sec:conclusion}.

\section{Related Work}
This section provides an overview of instance segmentation and few-shot learning.

\textbf{Instance segmentation} is the task of detecting objects in an image whilst also segmenting all the pixels that belong to them. Approaches generally fall into two categories: \textit{grouping-based} \cite{chen2018masklab,he2017mask,kuo2019shapemask,pinheiro2015learning} and \textit{proposal-based} \cite{bai2017deep,de2017semantic,kong2018recurrent, liu2017sgn} detection methods. The former employ a grouping strategy in which a network produces per-pixel information that is post-processed to obtain instance segmentations. In proposal-based methods, a model first identifies potential areas and subsequently classifies and segments these regions. The most widely used two-step detection method is Mask R-CNN \cite{he2017mask}, which uses a Region Proposal Network (RPN) to propose detection regions which are passed to classification, localization, and mask predictor heads. However, these approaches do not perform well with small amounts of training data \cite{yan2019meta}.

\textbf{Few-shot learning} enables models to accommodate new classes for which little training data is available. Often, an episodic methodology \cite{vinyals2016matching} is used by providing \textit{query} items to be classified into $N$ classes and a \textit{support set} containing training examples of the $N$ classes. Approaches for few-shot learning can largely be split up in \textit{optimization-based} \cite{andrychowicz2016learning,finn2017model,ravi2016optimization} and \textit{metric-learning} \cite{chen2019closer,gidaris2018dynamic,koch2015siamese,snell2017prototypical,sung2018learning,vinyals2016matching}.

\textit{Optimization-based} methods train a \textit{meta-learner} from a series of tasks such that it is able to generate weights for a \textit{learner} which learns parameters for new tasks that have few training examples. The meta-learner is generally modeled as an optimization procedure \cite{andrychowicz2016learning,finn2017model} or is a separate network enhanced with memory \cite{ mishra2017simple,ravi2016optimization} that uses previous tasks as experience and is trained to produce a learner.

\textit{Metric-learning} methods learn a feature embedding such that objects from the same class are close in the embedding space and objects of different classes are far apart. Koch \etal \cite{koch2015siamese} employed a Siamese network \cite{bromley1994signature}, where the distance between query and support image embeddings is minimized if they are of the same class, and maximized otherwise. Matching Networks \cite{vinyals2016matching} compute the distance between every learned query and support embedding, while Prototypical Networks \cite{snell2017prototypical} compute per-class representatives. Relation Networks \cite{sung2018learning} learn both a distance function and an embedding. In contrast to previous methods that focus solely on the performance on the novel classes, Gidaris and Komodakis \cite{gidaris2018dynamic} focus on classifying both novel and base classes jointly using a softmax cosine-similarity classifier along with a weight generator for novel classes. Recently, Chen \etal \cite{chen2019closer} have shown that fine-tuning on the novel classes, which was largely ignored previously, generally performs better than episodic training. Finally, Qi \etal \cite{qi2018low} propose \textit{weight-imprinting} by adding novel class embeddings into an existing weight matrix, allowing incremental addition of classes without training.

\textbf{Few-shot object detection} extends few-shot learning to object detection. RepMet \cite{schwartz2018repmet} trains a metric-learning sub-network to encode the support set, while Kang \etal \cite{kang2019few} directly train a meta-learner on top of YOLOv2~\cite{redmon2017yolo9000}. Inspired by \cite{chen2019closer}, Wang \etal developed TFA \cite{wang2020frustratingly}, which achieves state-of-the-art in object detection with a two-stage approach. Instead of fine-tuning the entire network, TFA first trains Faster R-CNN \cite{ren2015faster} on the base classes and then only fine-tunes the predictor heads.

\textbf{Few-shot instance segmentation.} Few works have addressed FSIS \cite{fan2020fgn,michaelis2018one,yan2019meta}. Most approaches provide guidance to certain parts of the Mask R-CNN architecture to ensure the network is better informed of the novel classes. Both Meta R-CNN \cite{yan2019meta} and Siamese Mask R-CNN \cite{michaelis2018one} compute embeddings of the support set and combine these with the feature map produced by the network backbone. The combination is implemented through different operations such as subtraction \cite{michaelis2018one} to focus the network on specific image areas, or concatenation \cite{yan2019meta} to provide additional information at a certain stage. FGN \cite{fan2020fgn} guides the RPN, RoI detector and mask upsampling layers with the support set feature embeddings through similar operations.

\textbf{Incremental few-shot object detection} has been considered in ONCE \cite{perez2020incremental}, which uses CenterNet \cite{zhou2019objects} as a backbone to learn a class-agnostic feature extractor and a per-class code generator network for novel classes.

\textbf{Incremental few-shot instance segmentation.} To our knowledge, we are the first to target incremental FSIS. FGN and Siamese Mask R-CNN depend on being passed examples of every class at test time, which requires a large amount of memory when considering many classes. Meta R-CNN can pre-compute per-class attention vectors, but requires retraining to handle a different number of classes. In contrast, our method can incrementally add classes without retraining or requiring examples of base classes. 


\begin{figure}[htb]
	\begin{center}
		\includegraphics[width=1.0\linewidth]{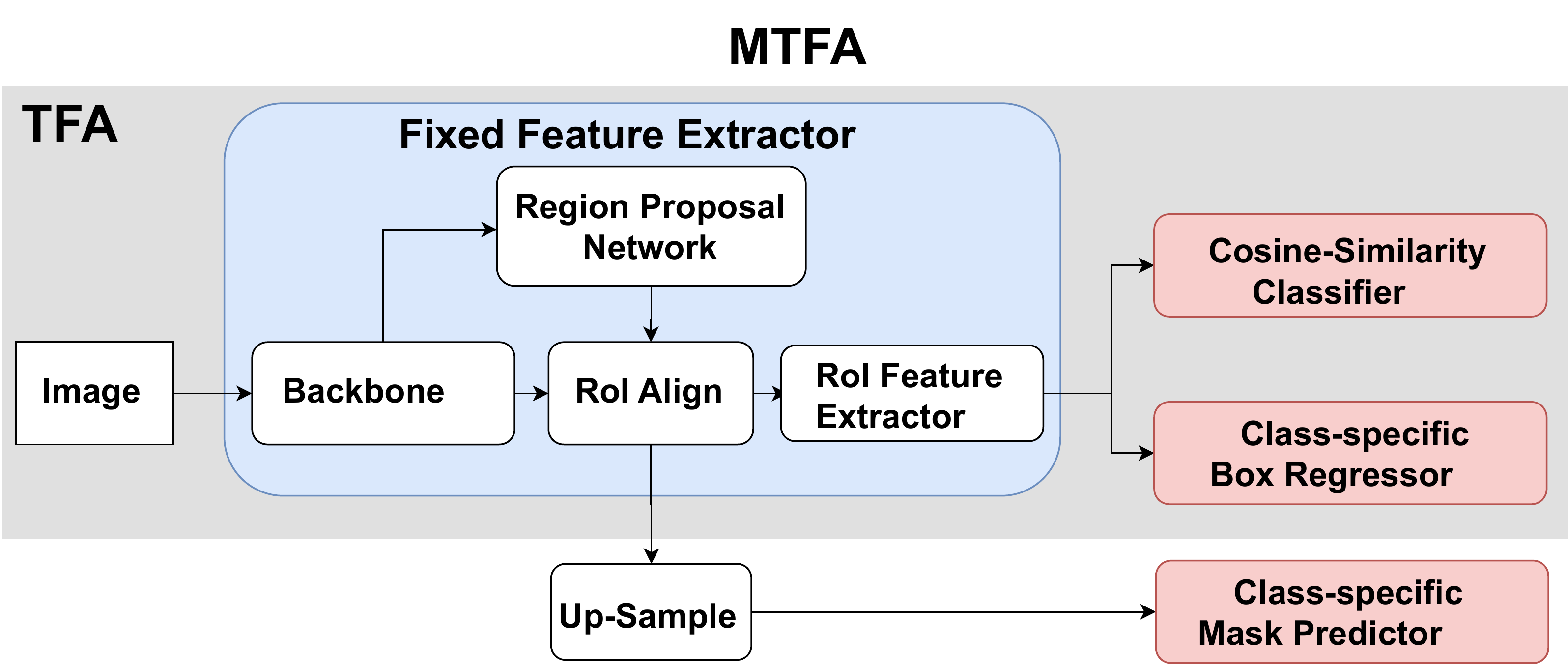}
	\end{center}
	\caption{\textbf{Architecture of TFA and MTFA.} MTFA extends TFA with a mask prediction branch. In the first training stage, the whole network is trained on the base classes. In the second stage, the feature extractor is frozen (blue) while the classifier and box and mask heads (in red) are fine-tuned on base and novel classes.\vspace{-2mm}}
	\label{fig:mtfa}
\end{figure}

\begin{figure*}[htb]
	\begin{center}
		\includegraphics[width=0.80\linewidth]{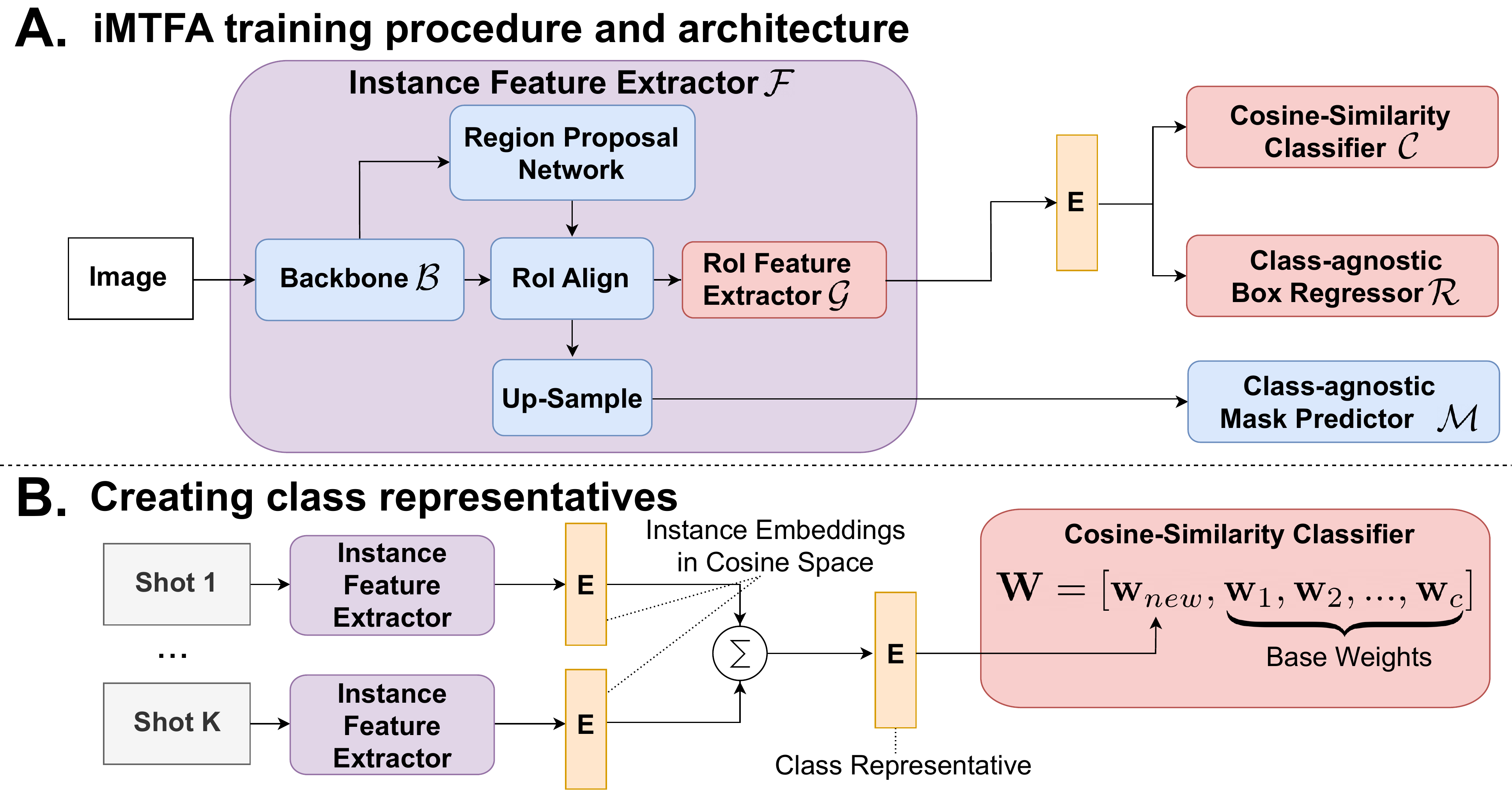}
		\vspace{-3mm}
	\end{center}
	\caption{\textbf{Architecture of iMTFA.} (A.) In the first stage, the whole network is trained. In the second stage, the blue components are frozen, while the RoI Feature Extractor $\mathcal{G}$ is trained to produce discriminative embeddings aligned with the class-representatives in the cosine-similarity classifier $\mathcal{C}$. Both stages are trained only on the base classes. (B.) Given the $K$ shots for each novel class, the IFE computes class weight vectors that are placed alongside the weights for the base classes.\vspace{-2mm}}
	\label{fig:big-image-method}
    \vspace{-1mm}
\end{figure*}

\section{Methodology} \label{sec:method}

We first introduce common terminology in few-shot learning (Section~\ref{sec:form_fewshot}). We then introduce our baseline few-shot instance segmentation method MTFA (Section~\ref{sec:mtfa}). In Section~\ref{iMTFASection}, we introduce our incremental method: iMTFA.

\subsection{Formulation of few-shot instance segmentation} \label{sec:form_fewshot}
In few-shot learning, we have a set of base classes $C_{base}$, for which a large amount of training data is available, and a disjoint set of novel classes $C_{novel}$, which has a small amount of training data. The goal is to train a model that does well on the novel classes $C_{test} = C_{novel}$ \cite{snell2017prototypical, vinyals2016matching} or on both base and novel classes jointly $C_{test} = C_{base} \cup C_{novel}$ \cite{gidaris2018dynamic}. 
In few-shot classification, Vinyals \etal \cite{vinyals2016matching} introduce the \textit{episodic-training} methodology. Episodic-training sets up a series of episodes $E_i = (\matr{I}^q, S_i)$ where $S_i$ is a support set containing $N$ classes from $C_{train} = C_{novel} \cup C_{base}$ along with $K$ examples per class (\textit{$N$-way $K$-shot}). A network is then tasked to classify an image $\matr{I}^q$, termed \textit{query}, out of the classes in $S_i$. The idea is that solving a different classification task each episode leads to better generalization and results on $C_{novel}$. This approach has also been extended to FSOD (e.g., \cite{kang2019few}) and FSIS (e.g., \cite{fan2020fgn, yan2019meta}) by considering all objects in an image as queries and having a single support set per-image instead of per-query.

The challenge of FSIS is not only to classify the query objects, but also to determine their localization and segmentation. Given a query image $\matr{I}^q$, FSIS produces labels $y_i$, bounding boxes $b_i$ and segmentation masks $\matr{M}_i$ for all objects in $\matr{I}^q$ that belong to $C_{test}$.

\subsection{MTFA: A non-incremental baseline approach} \label{sec:mtfa}
Our non-incremental baseline approach extends the Two-Stage Fine-tuning (TFA, \cite{wang2020frustratingly}) object-detection method introduced by Wan \etal. We first give an overview of TFA and then describe our extension, Mask-TFA (MTFA), which includes an instance segmentation task. In Section~\ref{iMTFASection}, we extend MTFA to an incremental approach.

\textbf{TFA} (Figure~\ref{fig:mtfa}) uses Faster R-CNN \cite{ren2015faster} with a two-stage training scheme. In the first stage, the network is trained on the base classes $C_{base}$. In the second stage, feature-extractor $\mathcal{F}$ is frozen and only the prediction heads are trained. $\mathcal{F}$ consists of network backbone $\mathcal{B}$, region proposal network (RPN) and RoI feature extractor $\mathcal{G}$. Thus, only RoI classifier $\mathcal{C}$ and box regressor $\mathcal{R}$ are fine-tuned in the second stage. Fine-tuning is performed on a dataset containing an equal number of examples of $C_{base}$ and $C_{novel}$ classes.

\textbf{MTFA.} We extend TFA similarly to how Mask R-CNN extends Faster R-CNN: by adding a mask prediction branch at the RoI level (Figure~\ref{fig:mtfa}). Thus, MTFA includes a branch with an up-sampling component and a mask predictor $\mathcal{M}$. We also employ a two-stage fine-tuning approach by first training the network on the base classes and then fine-tuning all predictor heads $\mathcal{C}$, $\mathcal{R}$ and $\mathcal{M}$ on a balanced dataset of $K$ shots for every class.

\textbf{Cosine-similarity classifier.} Similar to TFA and other recent metric-learning methods \cite{chen2019closer,gidaris2018dynamic}, a cosine-similarity classifier is used for $\mathcal{C}$ to learn more discriminative per-class representatives. $\mathcal{C}$ is a fully-connected layer which, given embeddings computed by the fixed feature extractor $\mathcal{F}$ for a RoI, produces classification scores $\matr{S}$. $\mathcal{C}$ is parameterized by weight matrix $\matr{W} \in \mathbb{R}^{e \times c}$ where $e$ is the size of an embedding vector produced by $\mathcal{F}$ and $c$ is the number of classes. We denote the columns of $\matr{W}$ as $\matr{w}_j \in \mathbb{R}^e$ such that $\matr{W} = [\matr{w}_1, \matr{w}_2, \ldots, \matr{w}_c]$. Similar to \cite{he2017mask, krizhevsky2012imagenet, yan2019meta}, classification scores $\matr{S}_{i,j}$ for the $i$-th object proposal of an image $\matr{X}$ and the $j$-th class are produced as:

\begin{equation} \label{eq:initial_score_eq}
	\matr{S}_{i,j} = \mathcal{F}(\matr{X})_i^{\top} \cdot \matr{w}_j.
\end{equation}

Normalizing both the output of the feature extractor $\mathcal{F}$ and the weights $\matr{w}_i$ causes $\mathcal{C}$ to compute the cosine similarity between $\mathcal{F}(\matr{X})_i$ and class-representative $\matr{w}_j$:

\begin{equation} \label{alpha_equation}
\matr{S}_{i,j} = \frac{\alpha \mathcal{F}(\matr{X})_i^{\top} \cdot \matr{w}_j}{\left\|\mathcal{F}(\matr{X})_i\right\|\left\|\matr{w}_j\right\|} ,
\end{equation}

where $\alpha$ is used to scale the scores before they are passed to a softmax layer.

Forcing all embeddings to align to a single class representative $\matr{w}_j$ results in class prototypes that are similar to prototypical networks \cite{snell2017prototypical}. Normalization bounds the dot product, which simplifies the network's training task by allowing only angular degrees of freedom for optimization.

{
\begin{figure*}[ht]
	\centering
	\footnotesize
	\setlength{\tabcolsep}{0.1em}
	\adjustbox{width=.9\linewidth}{
		\begin{tabular}{p{4mm}ccccc}
		    \rotatebox{90}{\hspace{4mm}\textbf{Success}} & 
			\includegraphics[width=1in, height=0.7in]{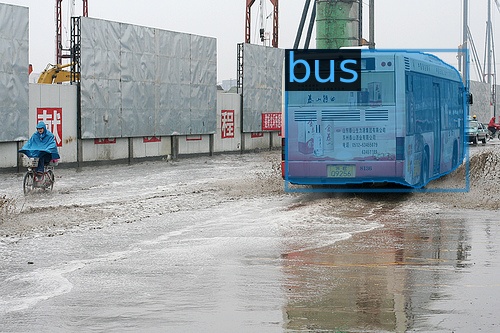} & \includegraphics[width=1in, height=0.7in]{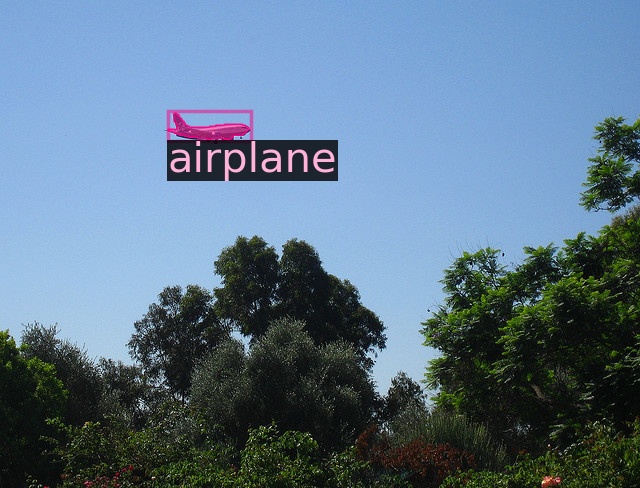} & \includegraphics[width=1in, height=0.7in]{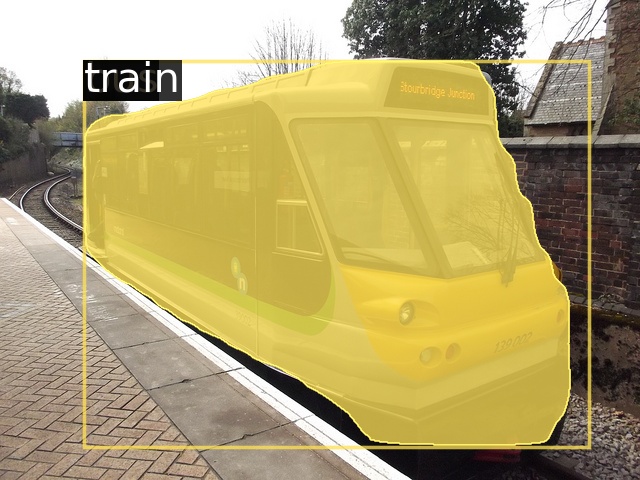} & \includegraphics[width=1in, height=0.7in]{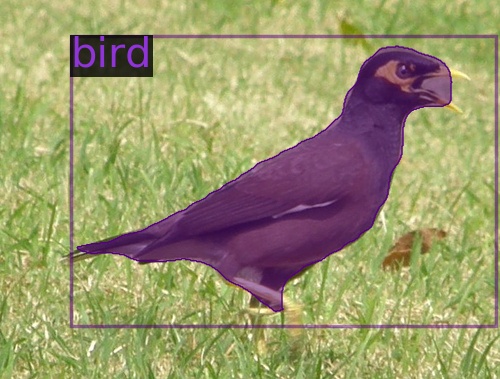} & \includegraphics[width=1in, height=0.7in]{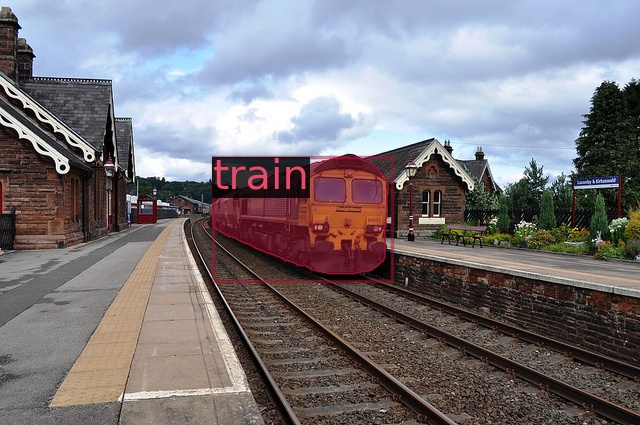} \\
			\multirow{2}{*}{\rotatebox{90}{\textbf{Failure}}} &
			\includegraphics[width=1in, height=0.7in]{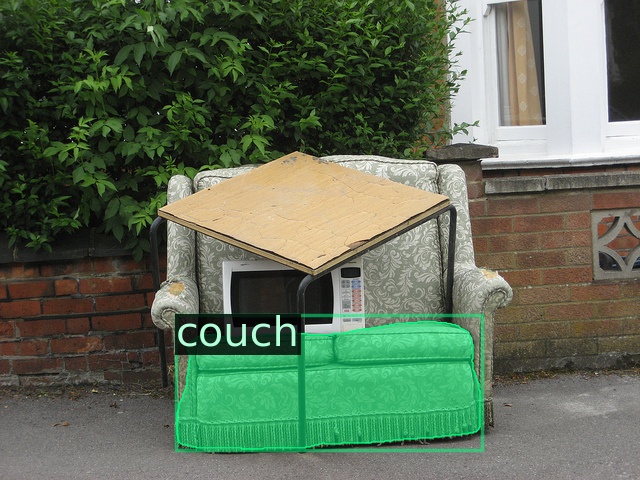} & \includegraphics[width=1in, height=0.7in]{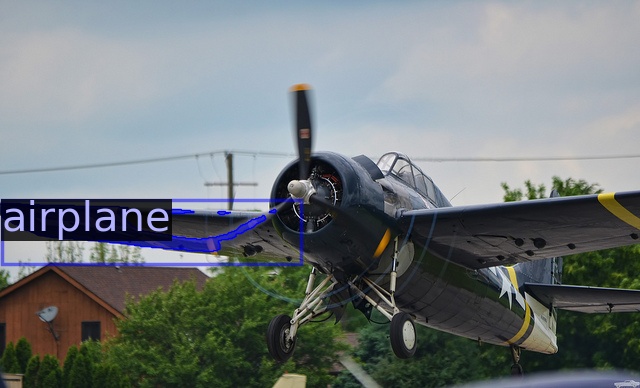} & \includegraphics[width=1in, height=0.7in]{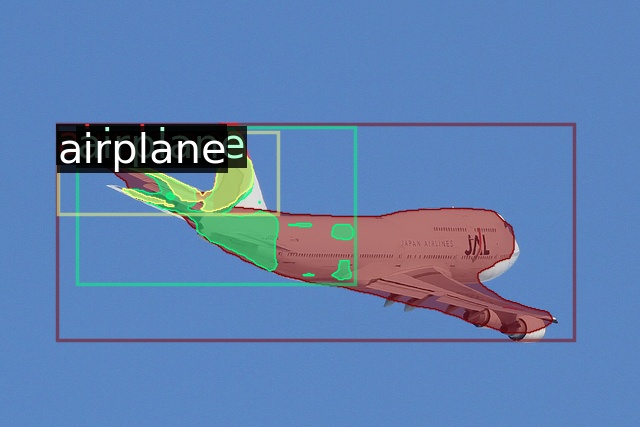} & \includegraphics[width=1in, height=0.7in]{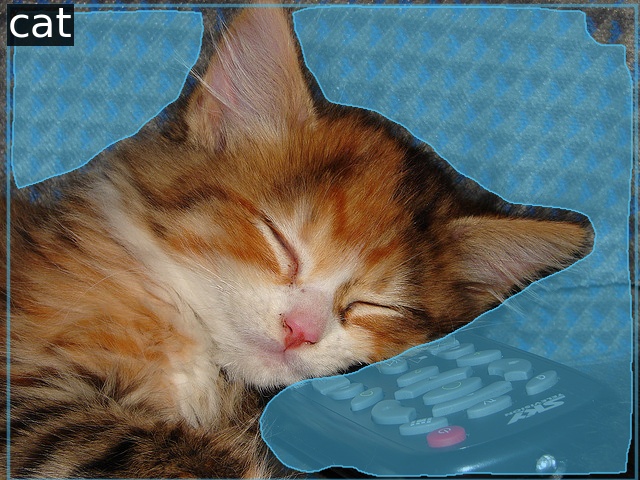} & \includegraphics[width=1in, height=0.7in]{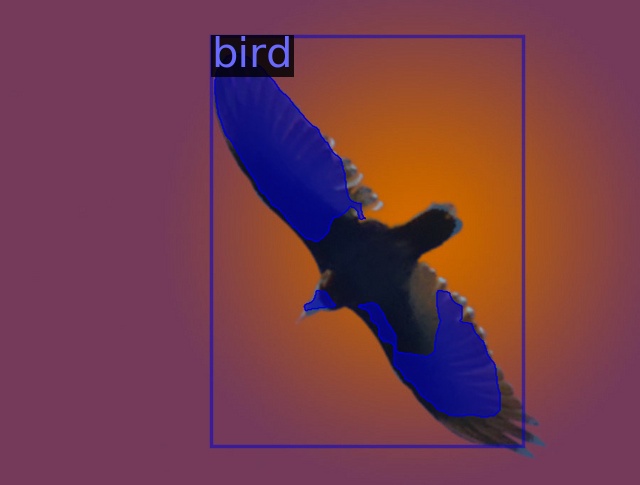} \\
			& 
			\includegraphics[width=1in, height=0.7in]{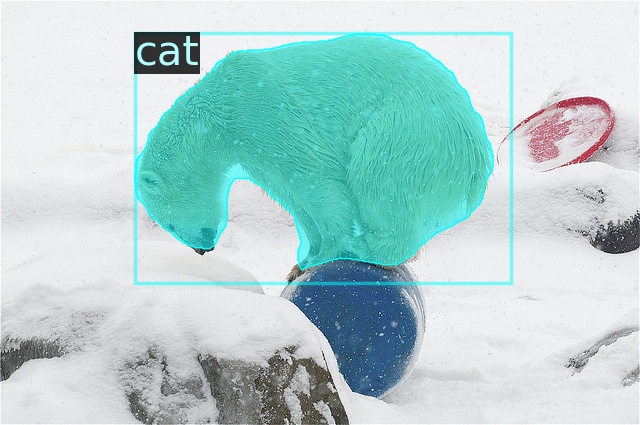} & \includegraphics[width=1in, height=0.7in]{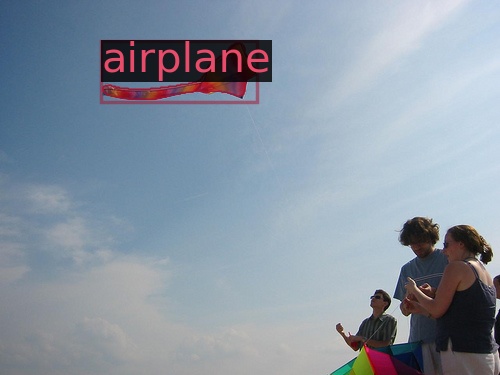} & \includegraphics[width=1in, height=0.7in]{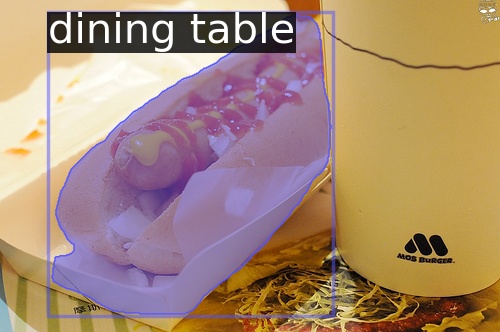} & \includegraphics[width=1in, height=0.7in]{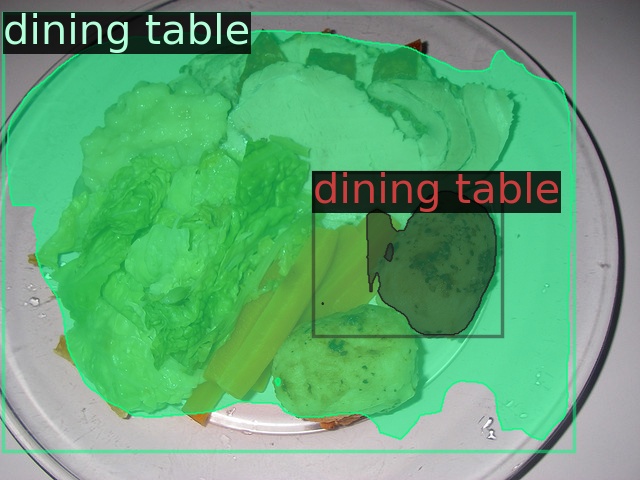} & \includegraphics[width=1in, height=0.7in]{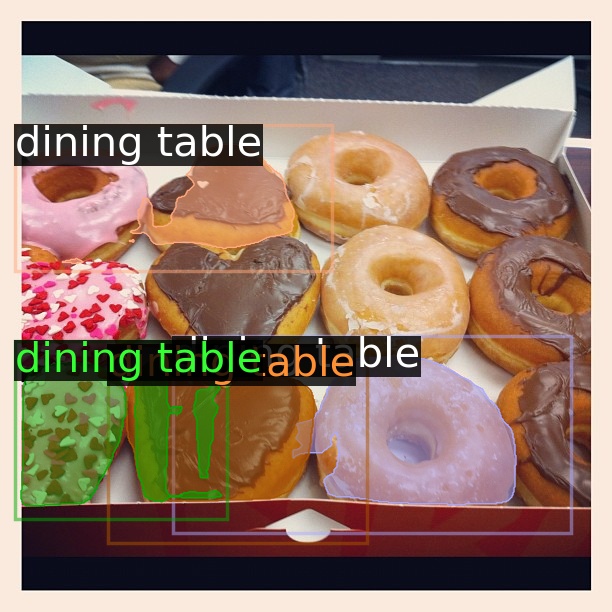} \\
	\end{tabular}}
    \caption{\textbf{Inference examples.} Successful (top row) and failure cases (bottom two rows), obtained on the 5-shot setting for the COCO novel classes. Failures include wrong classifications, wrong detections and inaccurate instance segmentations. See Section~\ref{reslabel} for details.}
    \label{fig:det-vis}
    \vspace{-3mm}
\end{figure*}}

\subsection{iMTFA: Incremental MTFA} \label{iMTFASection}
The main drawback of MTFA is the procedure of adding new classes. The second fine-tuning stage fixes the number of novel classes that can be recognized. Adding new classes requires this stage to be run again, which is not practical. The class-specific mask and box regressor heads also require adaptation to novel classes in the form of weights learned through fine-tuning. In this section, we extend MTFA to an incremental approach: iMTFA. To this end, we make the model class-agnostic and learn discriminative embeddings at the feature extractor level. These embeddings are used as novel class representatives in the classification head without further need for training. The architecture and procedure for adding new classes are depicted in Figure~\ref{fig:big-image-method}.

\textbf{Instance Feature Extractor (IFE)}. The fixed feature extractor $\mathcal{F}$ employed by TFA and MTFA is not trained to produce discriminative vector embeddings. Instead, the classification head $\mathcal{C}$ is fine-tuned in order to align the learned per-class weights $\matr{w}_i$ to fixed features computed by $\mathcal{F}$ for every RoI. We replace the fixed feature extractor by an instance feature extractor (IFE).

The key idea of our approach is to generate discriminative embeddings for each instance. To add new classes, the average of the generated embeddings is used as a per-class representative $\matr{w}_i$ in $\matr{W}$. This allows us to directly use instance embeddings as class representatives, without the need for fine-tuning.

The backbone $\mathcal{B}$ of Mask R-CNN produces feature maps for every RoI $\matr{R}_i = \mathcal{B}(\matr{X})_i$ with $i$ indicating one RoI. The RoI-level feature extractor $\mathcal{G}$, typically consisting of two fully-connected layers, then takes these feature maps and computes embeddings $\matr{z}_i$ that are compared to the per-class representatives $\matr{w}_i$ in the classifier head $\mathcal{C}$. The vector embedding for each RoI is thus:

\begin{equation}
\matr{z}_i = \mathcal{F}(\matr{X})_i = \mathcal{G}(\matr{R}_i) = \mathcal{G}(\mathcal{B}(\matr{X})_i)
\end{equation}

We propose to train $\mathcal{G}$ such that it produces discriminative embeddings per instance. This is achieved in two stages. First, we employ the same first training stage as MTFA -- fully training Mask R-CNN on the $C_{base}$ base classes. Second, we fine-tune $\mathcal{G}$ alongside the classifier $\mathcal{C}$ and box regressor $\mathcal{R}$, whilst keeping the backbone $\mathcal{B}$ and the RPN frozen. The fine-tuning is only performed on the set of base classes $C_{base}$, with the goal of generalizing to unseen classes in $C_{novel}$. 
The architecture of MTFA remains unchanged, only the training procedure is different. By training $\mathcal{G}$ as a sub-network with a cosine-similarity classifier, it produces embeddings which act as class representatives. 

\textbf{Creating class representatives.} The final goal is to create novel class weight vectors that can be placed alongside the weights for the base classes, held in $\mathcal{C}$'s weight matrix $\matr{W}$ after the second fine-tuning stage. To accomplish this, every image $\matr{X}$ containing one of the $K$ available novel shots is passed to the IFE, producing a feature embedding for each of the shots, $\matr{z}_i = \mathcal{F}(\matr{X})_i$. This is done for all $K$ shots, with novel class representatives $\matr{w}_{new}$ computed as:

\begin{equation}
\matr{w}_{new}= \frac{1}{K}\sum_{i=0}^{K} \frac{\matr{z}_i}{\left\|\matr{z}_i\right\|} .
\end{equation}

Because the normalized feature embeddings $\matr{z}_i$ are points on a hypersphere, their normalized average $\matr{w}_{new}$ is meaningful in this space and can be used as a class representative. We can pre-compute the class representatives and do not require all shots to be passed in at once. This greatly reduces the memory bottleneck in \cite{fan2020fgn,yan2019meta}.

\textbf{Class-agnostic box and mask predictors.} iMTFA does not need class-specific weights for the box regressor and mask predictor $\mathcal{R}$ and $\mathcal{M}$. Instead, we use class-agnostic variants of these components and can add new classes by simply averaging their computed embeddings and placing them in the classification head's weight matrix $\matr{W}$. This also implies that we can train on novel classes without providing instance masks.

\textbf{Inference.} Because we predict the localization and segmentation components in a class-agnostic manner, class representative are all we need at test time. The lowest cosine distance between an RoI's embedding and the class representatives gives us the class predictions.

\textbf{Relation to other methods.} Related approaches  \cite{fan2020fgn,michaelis2018one,yan2019meta} rely on being passed examples of every class during training and at test time. This causes a memory bottleneck and forces some methods to only report evaluation results on the ground truth classes in an image \cite{fan2020fgn,michaelis2018one}, train for significant amounts of time to match classes in a pairwise manner \cite{michaelis2018one}, or to use greatly reduced image sizes \cite{yan2019meta}. In contrast, iMTFA uses weight-imprinting \cite{qi2018low, siam2019amp} to  keep an internal memory for the class representatives, and thus does not need this memory consumption at test or train time.


\begin{table*}[]
\centering
\footnotesize
\adjustbox{width=\linewidth}{%
\begin{tabular}{|c|c|c|d{2.2}d{2.2}|d{2.2}d{2.2}|d{2.2}d{2.2}|d{2.2}d{2.2}|d{2.2}d{2.2}|d{2.2}d{2.2}|d{2.2}d{2.2}|}
\hline
  \multirow{3}{*}{\textbf{Shots}} & 
  \multirow{3}{*}{\textbf{Inc.}} &
  \multirow{3}{*}{\textbf{Method}} &
                      \multicolumn{6}{c|}{\textbf{Detection}}                                                                          & \multicolumn{6}{c|}{\textbf{Segmentation}}                                                                       \\ \cline{4-15} 
& &                                                                       & \multicolumn{2}{c|}{\textbf{Overall}} & \multicolumn{2}{c|}{\textbf{Base}} & \multicolumn{2}{c|}{\textbf{Novel}} & \multicolumn{2}{c|}{\textbf{Overall}} & \multicolumn{2}{c|}{\textbf{Base}} & \multicolumn{2}{c|}{\textbf{Novel}} \\ \cline{4-15}
 &  &  & \multicolumn{1}{c}{\textbf{AP}}       & \multicolumn{1}{c|}{\textbf{AP50}}     & \multicolumn{1}{c}{\textbf{AP}}      & \multicolumn{1}{c|}{\textbf{AP50}}   & \multicolumn{1}{c}{\textbf{AP}}      & \multicolumn{1}{c|}{\textbf{AP50}}    & \multicolumn{1}{c}{\textbf{AP}}       & \multicolumn{1}{c|}{\textbf{AP50}}     & \multicolumn{1}{c}{\textbf{AP}}      & \multicolumn{1}{c|}{\textbf{AP50}}   & \multicolumn{1}{c}{\textbf{AP}}      & \multicolumn{1}{c|}{\textbf{AP50}}    \\ \hline
\multirow{4}{*}{1}                   & \multirow{2}{*}{}                 & Base-Only       &  \multicolumn{1}{B{.}{.}{2,2}}{28.67}   & \multicolumn{1}{B{.}{.}{2,2}|}{43.53}    & \multicolumn{1}{B{.}{.}{2,2}}{38.22}   & \multicolumn{1}{B{.}{.}{2,2}|}{58.04}  & \multicolumn{1}{c}{\textemdash}              & \multicolumn{1}{c|}{\textemdash}              & \multicolumn{1}{B{.}{.}{2,2}}{26.34}    & \multicolumn{1}{B{.}{.}{2,2}|}{41.55}    & \multicolumn{1}{B{.}{.}{2,2}}{35.12}   & \multicolumn{1}{B{.}{.}{2,2}|}{55.40}  & \multicolumn{1}{c}{\textemdash}              & \multicolumn{1}{c|}{\textemdash}              \\
                                     &                                     & MTFA            & 24.32             & 39.64             & 31.73            & 51.49           & 2.10              & 4.07             & 22.98             & 37.48             & 29.85            & 48.64           & 2.34             & 3.99             \\ \cline{2-15} 
                                     & \multirow{2}{*}{\checkmark}                & ONCE            & 13.6              & \multicolumn{1}{c|}{N/A}              & 17.9             & \multicolumn{1}{c|}{N/A}            & 0.7              & \multicolumn{1}{c|}{N/A}             & \multicolumn{1}{c}{\textemdash}               & \multicolumn{1}{c|}{\textemdash}               & \multicolumn{1}{c}{\textemdash}              & \multicolumn{1}{c|}{\textemdash}             & \multicolumn{1}{c}{\textemdash}              & \multicolumn{1}{c|}{\textemdash}              \\
                                     &                                     & iMTFA           & 21.67             & 31.55             & 27.81            & 40.11           & \multicolumn{1}{B{.}{.}{2,2}}{3.23}    & \multicolumn{1}{B{.}{.}{2,2}|}{5.89}    & 20.13             & 30.64             & 25.9             & 39.28           & \multicolumn{1}{B{.}{.}{2,2}}{2.81}    & \multicolumn{1}{B{.}{.}{2,2}|}{4.72}    \\ \hline
\multirow{4}{*}{5}                   & \multirow{2}{*}{}                 & Base-Only       & \multicolumn{1}{B{.}{.}{2,2}}{28.67}    & \multicolumn{1}{B{.}{.}{2,2}|}{43.53}    & \multicolumn{1}{B{.}{.}{2,2}}{38.22}   & \multicolumn{1}{B{.}{.}{2,2}|}{58.04}  & \multicolumn{1}{c}{\textemdash}              & \multicolumn{1}{c|}{\textemdash}              & \multicolumn{1}{B{.}{.}{2,2}}{26.34}    & \multicolumn{1}{B{.}{.}{2,2}|}{41.55}    & \multicolumn{1}{B{.}{.}{2,2}}{35.12}   & \multicolumn{1}{B{.}{.}{2,2}|}{55.40}  & \multicolumn{1}{c}{\textemdash}              & \multicolumn{1}{c|}{\textemdash}              \\
                                     &                                     & MTFA            & 26.39             & 41.52             & 33.11            & 51.49           & \multicolumn{1}{B{.}{.}{2,2}}{6.22}    & \multicolumn{1}{B{.}{.}{2,2}|}{11.63}   & 25.07             & 39.95             & 31.29            & 49.55           & \multicolumn{1}{B{.}{.}{2,2}}{6.38}    & \multicolumn{1}{B{.}{.}{2,2}|}{11.14}   \\ \cline{2-15} 
                                     & \multirow{2}{*}{\checkmark}                & ONCE            & 13.7              & \multicolumn{1}{c|}{N/A}              & 17.9             & \multicolumn{1}{c|}{N/A}            & 1.0              & \multicolumn{1}{c|}{N/A}             & \multicolumn{1}{c}{\textemdash}               & \multicolumn{1}{c|}{\textemdash}               & \multicolumn{1}{c}{\textemdash}              & \multicolumn{1}{c|}{\textemdash}             & \multicolumn{1}{c}{\textemdash}              & \multicolumn{1}{c|}{\textemdash}              \\
                                     &                                     & iMTFA           & 19.62             & 28.06             & 24.13            & 33.69           & 6.07             & 11.15            & 18.22             & 27.10              & 22.56            & 33.25           & 5.19             & 8.65             \\ \hline
\multirow{4}{*}{10}                  & \multirow{2}{*}{}                 & Base-Only       & \multicolumn{1}{B{.}{.}{2,2}}{28.67}    & \multicolumn{1}{B{.}{.}{2,2}|}{43.53}    & \multicolumn{1}{B{.}{.}{2,2}}{38.22}   & \multicolumn{1}{B{.}{.}{2,2}|}{58.04}  & \multicolumn{1}{c}{\textemdash}              & \multicolumn{1}{c|}{\textemdash}              & \multicolumn{1}{B{.}{.}{2,2}}{26.34}    & \multicolumn{1}{B{.}{.}{2,2}|}{41.55}    & \multicolumn{1}{B{.}{.}{2,2}}{35.12}   & \multicolumn{1}{B{.}{.}{2,2}|}{55.40}  & \multicolumn{1}{c}{\textemdash}              & \multicolumn{1}{c|}{\textemdash}              \\
                                     &                                     & MTFA            & 27.44             & 42.84             & 33.83            & 52.04           & \multicolumn{1}{B{.}{.}{2,2}}{8.28}    & \multicolumn{1}{B{.}{.}{2,2}|}{15.25}   & 25.97             & 41.28             & 31.84            & 50.17           & \multicolumn{1}{B{.}{.}{2,2}}{8.36}    & \multicolumn{1}{B{.}{.}{2,2}|}{14.58}   \\ \cline{2-15} 
                                     & \multirow{2}{*}{\checkmark}                & ONCE            & 13.7              & \multicolumn{1}{c|}{N/A}              & 17.9             & \multicolumn{1}{c|}{N/A}            & 1.2              & \multicolumn{1}{c|}{N/A}             & \multicolumn{1}{c}{\textemdash}               & \multicolumn{1}{c|}{\textemdash}               & \multicolumn{1}{c}{\textemdash}              & \multicolumn{1}{c|}{\textemdash}             & \multicolumn{1}{c}{\textemdash}              & \multicolumn{1}{c|}{\textemdash}              \\
                                     &                                     & iMTFA           & 19.26             & 27.49             & 23.36            & 32.41           & 6.97             & 12.72            & 17.87             & 26.46             & 21.87            & 32.01           & 5.88             & 9.81             \\ \hline
\end{tabular}
}
\caption{\textbf{FSOD and FSIS performance on COCO for both base and novel classes.} ONCE cannot perform instance segmentation and Base-Only does not consider novel classes. For ONCE, no AP50 is reported. Both MTFA and iMTFA outperform ONCE for object detection while also being able to perform instance segmentation. Inc. stands for incremental.\vspace{-3mm}} 
\label{tab:cocoall}
\end{table*}

\section{Experiments} \label{sec:experiments}

We first introduce our experiment setting (Section~\ref{subsec:experiment_setting}) and the implementation details (Section~\ref{subsec:implementation_details}). Then we evaluate iMTFA and MTFA and compare them to related approaches (Section~\ref{reslabel}), followed by an ablation study.

\subsection{Experiment setup} \label{subsec:experiment_setting}
Our main evaluation procedure follows conventions established in FSOD \cite{kang2019few,wang2020frustratingly,yan2019meta}. We evaluate on the  COCO \cite{lin2014microsoft}, VOC2007 \cite{everingham2010pascal} and VOC2012 \cite{everingham2010pascal} datasets. We split the 80 COCO classes as proposed in \cite{kang2019few}. The 20 classes that intersect with VOC are set as novel classes and the remaining 60 classes as base classes. The union of COCO's 80k train and 35k validation images are used for training and the remaining $\sim$5k images are the test set. The VOC dataset combines VOC2007 and VOC2012 and the resulting validation set is used for testing. We evaluate the performance of having $K = 1,5,10$ shots per novel class. To reduce the effect of outliers as a result of the random selection of the $K$ shots, we run all tests $10$ times with $K$ random examples per class, and report the mean result. Our few-shot evaluation procedure is the same as in \cite{wang2020frustratingly}.

\textbf{Comparison with other methods.} We compare the instance segmentation performance of iMTFA and MTFA with the three other known FSIS methods: Meta R-CNN \cite{yan2019meta}, Siamese Mask R-CNN \cite{michaelis2018one}, and FGN \cite{fan2020fgn}. Additionally, we compare the object detection performance to the only known incremental FSOD method, ONCE \cite{perez2020incremental}.

For Meta R-CNN and ONCE, we use the method described above. However, Siamese Mask R-CNN and FGN use an evaluation scheme in which the classes that belong to each query image are known during testing. Only the classes that appear in the ground truth are included in the support set for each image. This eliminates many potential false positives, since similar classes that do not appear in an image together will not be confused. In contrast, we perform a \textit{$N$-way $K$-shot} evaluation for every image, where $N$ is the number of test classes. This ensures that all classes in a dataset can be detected in an image. For comparison, we emulate an evaluation method with a similar property by zeroing the probabilities computed by our softmax classifier for classes that do not appear in the query image. This discredits our methods, since it leaves non-occurring classes inside our metric space. Nevertheless, it serves as a lower-bound for the performance of non-episodic testing. We name this procedure \textit{ground-truth only evaluation} (GTOE).

To compare against Siamese Mask R-CNN, we use one of their evaluation setups, which we term COCO-Split-2. This split consists of COCO classes with indices $4k, 1 \leq k \leq 20$ as novel classes, leaving the rest as base classes. The ResNet-50 backbone of Siamese Mask R-CNN is trained on 687 classes from ImageNet1K~\cite{russakovsky2015imagenet} that do not overlap with the classes in COCO. Unfortunately, it is unknown which classes have been used, so we opt to train on all 1,000 classes in ImageNet1K. To compare against FGN, we use the COCO2VOC setup, where we train on COCO but test on the VOC test set.

Following COCO evaluation practices, we report the performance using AP and AP50, using an intersection-over-union (IoU) overlap of bounding boxes and masks, for object detection and instance segmentation, respectively.

\subsection{Implementation details} \label{subsec:implementation_details}
Our Mask R-CNN \cite{he2017mask} is implemented using Detectron2 \cite{wu2019detectron2}. Our backbone is a ResNet-50 \cite{he2016deep} with a Feature Pyramid Network \cite{lin2017feature}. All models are trained using SGD and a batch size of 8 on two NVIDIA V100s, with four images per GPU. The second fine-tuning stage has a learning rate of 0.0007 for iMTFA and a learning rate of 0.0005 for MTFA. We set the cosine-similarity scaling factor $\alpha$ to 1.0 for the iMTFA COCO-Novel, 10.0 for iMTFA COCO-All and 20.0 for MTFA (see also Section~\ref{ablationsec}). Mask R-CNN has many parameters, hence we encourage the reader to visit the public repository for more details.

\begin{table}[ht]
	\centering
	\caption{\textbf{FSOD and FSIS performance on the COCO novel classes.} MTFA and iMTFA outperform the current state-of-the-art in terms of AP. Inc. stands for incremental.\vspace{-3mm}
	}
    \resizebox{\columnwidth}{!}{%
		\begin{tabular}{|c|c|l|d{2.2}d{2.2}|d{2.2}d{2.2}|}
\hline
  \multirow{2}{*}{\textbf{\#}} & 
  \multirow{2}{*}{\textbf{Inc.}} &
  \multirow{2}{*}{\textbf{Method}} &
  \multicolumn{2}{c|}{\textbf{Detection}} &
  \multicolumn{2}{c|}{\textbf{Segmentation}} \\
  &
  &
  &
  \multicolumn{1}{c}{\textbf{AP}} &
  \multicolumn{1}{c|}{\textbf{AP50}} &
  \multicolumn{1}{c}{\textbf{AP}} &
  \multicolumn{1}{c|}{\textbf{AP50}} \\ \hline
\multirow{2}{*}{1}  & \multicolumn{1}{c|}{}                  & MTFA         & 2.47 & 4.85  & 2.66 & 4.56  \\ \cline{2-7} 
                    & \multicolumn{1}{c|}{\checkmark}                 & iMTFA        & \multicolumn{1}{B{.}{.}{2,2}}{3.28} & \multicolumn{1}{B{.}{.}{2,2}|}{6.01}  & \multicolumn{1}{B{.}{.}{2,2}}{2.83} & \multicolumn{1}{B{.}{.}{2,2}|}{4.75}  \\ \hline
\multirow{4}{*}{5}  & \multicolumn{1}{c|}{\multirow{3}{*}{}} & MRCN+ft-full & 1.3 & 3.0  & 1.3 & 2.7  \\
                    & \multicolumn{1}{c|}{}                    & Meta R-CNN   & 3.5 & 9.9  & 2.8 & 6.9  \\
                    & \multicolumn{1}{c|}{}                    & MTFA         & \multicolumn{1}{B{.}{.}{2,2}}{6.61} & \multicolumn{1}{B{.}{.}{2,2}|}{12.32} &\multicolumn{1}{B{.}{.}{2,2}}{6.62} & \multicolumn{1}{B{.}{.}{2,2}|}{11.58} \\ \cline{2-7} 
                    & \multicolumn{1}{c|}{\checkmark}                 & iMTFA        & 6.22 & 11.28 & 5.24 & 8.73  \\ \hline
\multirow{4}{*}{10} & \multicolumn{1}{c|}{\multirow{3}{*}{}} & MRCN+FT-full & 2.5 & 5.7  & 1.9 & 4.7  \\
                    & \multicolumn{1}{c|}{}                    & Meta R-CNN   & 5.6 & 14.2 & 4.4 & 10.6 \\
                    & \multicolumn{1}{c|}{}                    & MTFA         & \multicolumn{1}{B{.}{.}{2,2}}{8.52} &\multicolumn{1}{B{.}{.}{2,2}|}{ 15.53} & \multicolumn{1}{B{.}{.}{2,2}}{8.39} &\multicolumn{1}{B{.}{.}{2,2}|}{ 14.64} \\ \cline{2-7} 
                    & \multicolumn{1}{c|}{\checkmark}                 & iMTFA        & 7.14 & 12.91 & 5.94 & 9.96 \\ \hline
\end{tabular}}
	\label{tab:novel-COCO}
\end{table}

\subsection{Results} \label{reslabel}
\textbf{Results on the COCO novel classes.} We compare against Meta R-CNN \cite{yan2019meta} and a fully-converged Mask R-CNN model fine-tuned on the novel classes (MRCN+ft-full, \cite{yan2019meta}). We report object detection and instance segmentation performance on the 20 COCO novel classes (COCO-Novel) in Table~\ref{tab:novel-COCO}. For all methods, detection and segmentation performance increases with the number of shots. Both iMTFA and MTFA outperform Meta R-CNN and MRCN+ft-full by a large margin in terms of AP, for every number of tested shots. In terms of AP50, MTFA surpasses Meta R-CNN but iMTFA is slightly behind. This suggests we may have difficulties finding the coarse location of an object but perform better at higher IoU thresholds.

Meta R-CNN directly uses image crops of the $K$ available shots per class to infer class-attentive vectors. In contrast, iMTFA and MTFA re-use the largest part of the network $\mathcal{F}$ by directly working at a feature-map level. We argue this generates more representative vector embeddings for the novel shots, which would explain the large performance gap between Meta R-CNN and our methods. 

Examples of inference results for iMTFA on the COCO novel classes with $K = 5$ appear in Figure~\ref{fig:det-vis}. Successful segmentations are generally accurate. Failure cases include correctly classifying but incorrectly localizing an object (row 2, columns 1--3), correctly classifying and localizing but incorrectly segmenting (row 2, columns 4--5), and incorrectly classifying but correctly localizing and segmenting (row 3, columns 1--2). Classes that are diverse in appearance have more false positives. This is noticeable especially for the \texttt{dining table} class. Many objects that resemble food will be incorrectly classified as a dining table. For the \texttt{person} class, a similar trend is observed.

\textbf{Results on both base and novel COCO classes.} In this experiment, we strive to detect all 80 COCO classes (COCO-All). We report the standard evaluation of AP and AP50 over the 80 COCO classes. Additionally, we report the performance of the base and novel classes individually. We are the first to report performance for $C_{test} = C_{base} \cup C_{novel}$ in FSIS. To understand the merits of our approaches, we compare the object detection performance of iMTFA and MTFA with the state-of-the-art incremental FSOD method ONCE \cite{perez2020incremental}. We also report on a model trained only on the base classes (Base-Only). While this model cannot be used for the novel classes, it demonstrates how much the performance on base classes is affected.

Results are summarized in Table~\ref{tab:cocoall}. As expected, Base-Only performs best on the base classes. iMTFA surpasses the object detection performance of ONCE in terms of base and novel class performance. MTFA consistently outperforms iMTFA on the base classes, which may be caused by iMTFA's inability to adapt to existing per-class representatives when generating new ones. See also Section~\ref{ablationsec}. Apart from $K=1$, MTFA also performs better than iMTFA on the novel classes, in line with the results on COCO-All.

\begin{table}[htb]
    \centering
	\caption{\textbf{FSIS performance on COCO-Split-2.} iMTFA outperforms Siamese Mask R-CNN for $K=1$ and $K=5$, while MTFA performs best on $K=5$.\vspace{-3mm}}
	\label{tab:like-oneshot}
    \resizebox{\columnwidth}{!}{%
\begin{tabular}{|c|c|l|d{2.2}d{2.2}|d{2.2}d{2.2}|}
\hline
\multirow{2}{*}{\textbf{\#}} & 
  \multirow{2}{*}{\textbf{Inc.}} &
  \multirow{2}{*}{\textbf{Method}} &
  \multicolumn{2}{c|}{\textbf{Detection}} &
  \multicolumn{2}{c|}{\textbf{Segmentation}} \\
 &  &  & \multicolumn{1}{c}{\textbf{AP}} & \multicolumn{1}{c|}{\textbf{AP50}} & \multicolumn{1}{c}{\textbf{AP}} & \multicolumn{1}{c|}{\textbf{AP50}} \\ \hline
\multicolumn{1}{|c|}{\multirow{3}{*}{1}} &
  \multicolumn{1}{c|}{\multirow{2}{*}{}} &
  Siamese Mask R-CNN &
  8.6 &
  15.3 &
  6.7 &
  13.5 \\
\multicolumn{1}{|c|}{} &
  \multicolumn{1}{c|}{} &
  MTFA &
  8.26 &
  15.24 &
  8.25 &
  14.31 \\ \cline{2-7} 
\multicolumn{1}{|c|}{} &
  \multicolumn{1}{c|}{\checkmark} &
  iMTFA &
  \multicolumn{1}{B{.}{.}{2,2}}{10.06} &
  \multicolumn{1}{B{.}{.}{2,2}|}{17.85} &
  \multicolumn{1}{B{.}{.}{2,2}}{8.67} &
  \multicolumn{1}{B{.}{.}{2,2}|}{15.47} \\ \hline
\multicolumn{1}{|c|}{\multirow{3}{*}{5}} &
  \multicolumn{1}{c|}{\multirow{2}{*}{}} &
  Siamese Mask R-CNN &
  9.4 &
  16.8 &
  7.4 &
  14.8 \\
\multicolumn{1}{|c|}{} &
  \multicolumn{1}{c|}{} &
  MTFA &
  \multicolumn{1}{B{.}{.}{2,2}}{15.80} &
  \multicolumn{1}{B{.}{.}{2,2}|}{28.12} &
  \multicolumn{1}{B{.}{.}{2,2}}{15.14} &
  \multicolumn{1}{B{.}{.}{2,2}|}{25.83} \\ \cline{2-7} 
\multicolumn{1}{|c|}{} &
  \multicolumn{1}{c|}{\checkmark} &
  iMTFA &
  14.55 &
  25.73 &
  12.33 &
  21.95 \\ \hline
\end{tabular}%
}
\end{table}

\textbf{Comparison with Siamese Mask R-CNN}
We follow the GTOE evaluation procedure described in Section~\ref{subsec:experiment_setting} and report AP and AP50 for COCO-Split-2 in Table~\ref{tab:like-oneshot}. 

For both object detection and instance segmentation, MTFA and iMTFA outperform Siamese Mask R-CNN. Siamese Mask R-CNN uses image crops for the $K$ shots to guide the network. This may prove to be detrimental in terms of performance, similar to Meta R-CNN. Additionally, the learned embeddings may not have strong discriminative power since they are not directly optimized through the loss function. In contrast, iMTFA's IFE is trained to produce discriminative embeddings for the $K$ shots.

Our higher performance might also be due to our use of a cosine-similarity classifier, which has been shown to produce more meaningful embeddings than the binary cross-entropy loss employed by Siamese Mask R-CNN \cite{wojke2018deep}. Finally, our models are trained on all 1,000 ImageNet1K classes, whereas Siamese Mask R-CNN only uses 687.

\begin{table}[htb]
    \centering
	\caption{\textbf{FSIS performance on COCO2VOC.} For FGN, no AP results are reported by the authors.\vspace{-1mm}}
	\label{tab:like-fgn}
    \adjustbox{width=0.95\linewidth}{%
\begin{tabular}{|c|c|l|d{2.2}d{2.2}|d{2.2}d{2.2}|}
\hline
\multirow{2}{*}{\textbf{\#}} &
  \multirow{2}{*}{\textbf{Inc.}} &
  \multirow{2}{*}{\textbf{Method}} &
  \multicolumn{2}{c|}{\textbf{Detection}} &
  \multicolumn{2}{c|}{\textbf{Segmentation}} \\ \cline{4-7} 
                   &                   &       & \multicolumn{1}{c}{\textbf{AP}} & \multicolumn{1}{c|}{\textbf{AP50}} & \multicolumn{1}{c}{\textbf{AP}} & \multicolumn{1}{c|}{\textbf{AP50}} \\ \hline
\multirow{3}{*}{1} & \multirow{2}{*}{} & FGN   & \multicolumn{1}{c}{N/A}                              & \multicolumn{1}{B{.}{.}{2,2}|}{30.8}      & \multicolumn{1}{c}{N/A}         & 16.2                               \\  
                   &                   & MTFA  & 9.99                             & 21.68         & \multicolumn{1}{B{.}{.}{2,2}}{9.51}    & \multicolumn{1}{B{.}{.}{2,2}|}{19.28}                          \\ \cline{2-7} 
                   & \checkmark                 & iMTFA & \multicolumn{1}{B{.}{.}{2,2}}{11.47}   & 22.41         & 8.57        & 16.32                              \\ \hline
\end{tabular}}
\end{table}

\textbf{Comparison with FGN}. We compare iMTFA and MTFA to FGN using the cross-dataset COCO2VOC evaluation setting and the GTOE evaluation procedure. The FGN paper evaluates 1-way 1-shot, 3-way 1-shot and 3-way 3-shot performance. Since FGN's source code is not released and the used evaluation scheme is not common, we are only able to compare against 1-way 1-shot results. From Table~\ref{tab:like-fgn}, it shows that MTFA has superior performance in terms of instance segmentation while iMTFA's incremental approach is on par with FGN.

FGN's higher object detection performance suggests that guidance at the RPN and classifier stages is effective, although the better performance could partly be due to the use of a deeper backbone (ResNet-101 vs. ResNet-50). A combined approach appears promising for future work. Although the instance segmentation performance between iMTFA and FGN is similar, iMTFA  maintains the key advantage of being incremental.

\begin{table}[htb]
\centering
\adjustbox{width=1.0\linewidth}{%
\begin{tabular}{|c|c|l|d{2.2}d{2.2}|d{2.2}d{2.2}|}
\hline
\multirow{2}{*}{\textbf{\#}} & 
  \multirow{2}{*}{\textbf{Inc.}} &
  \multirow{2}{*}{\textbf{Method}} &
  \multicolumn{2}{c|}{\textbf{Detection}} &
  \multicolumn{2}{c|}{\textbf{Segmentation}} \\ \cline{4-7}
  &
  &
 &
  \multicolumn{1}{c}{\textbf{AP}} &
  \multicolumn{1}{c|}{\textbf{AP50}} &
  \multicolumn{1}{c}{\textbf{AP}} &
  \multicolumn{1}{c|}{\textbf{AP50}} \\ \hline
\multirow{4}{*}{5} & \multirow{3}{*}{} & MTFA                               & 6.61          & 12.32          & \multicolumn{1}{B{.}{.}{2,2}}{6.62} & \multicolumn{1}{B{.}{.}{2,2}|}{11.58} \\
& & CA MTFA                & \multicolumn{1}{B{.}{.}{2,2}}{7.00} & \multicolumn{1}{B{.}{.}{2,2}|}{12.58} & 6.11          & 10.16          \\
& & CA MTFA w/o FT $\mathcal{M}$ & 7.00          & 12.64          & 5.83          & 9.48           \\ \cline{2-7}
& \checkmark & iMTFA                              & 6.22          & 11.28          & 5.24          & 8.73           \\ \hline
\end{tabular}%
}
\caption{\textbf{Ablation MTFA/iMTFA.} Comparison between different variants of MTFA and iMTFA on COCO-Novel.\vspace{-3mm}}
\label{tab:compareimtfaandmtfa}
\end{table}

\subsection{Ablation study} \label{ablationsec}
We perform several ablations on the COCO 5-shot setting for novel classes.

\textbf{Comparison between iMTFA and MTFA.} \label{experiment:imtfavsMTFA} We identify two main reasons that can account for the performance difference between MTFA and iMTFA: using class-specific components and adjusting parts of the network through fine-tuning. To measure their effect, we compare MTFA and iMTFA along with (1) MTFA with a class-agnostic mask predictor $\mathcal{M}$ and box regressor $\mathcal{R}$ (CA MTFA) and (2) a class-agnostic MTFA without fine-tuning the mask predictor $\mathcal{M}$ (CA MTFA w/o FT $\mathcal{M}$). Results appear in Table~\ref{tab:compareimtfaandmtfa}.

Class-specific components and fine-tuning both help MTFA to achieve better segmentation performance. iMTFA is unable to adjust the generated novel weights based on existing weights in the metric space, which fine-tuning can do. We also find that fine-tuning the class-agnostic mask predictor is advantageous. This may be because iMTFA does not explicitly use the segmentation information for the $K$ shots to inform the mask predictor, whereas MTFA and class-agnostic MTFA achieve this by optimizing the segmentation loss directly. The performance loss from class-specific to class-agnostic is in line with \cite{he2017mask} and may be attributed to the additional number of trainable parameters.  

\begin{table}[htb]
    \centering
	\caption{\textbf{Ablation second fine-tuning stage.} Results on COCO-Novel for different classification heads in iMTFA.}
	\label{tab:extrafinetuning}
\adjustbox{width=1.0\linewidth}{%
\begin{tabular}{|c|c|l|d{2.2}d{2.2}|d{2.2}d{2.2}|}
\hline
\multirow{2}{*}{\textbf{\#}} & 
  \multirow{2}{*}{\textbf{Inc.}} &
  \multirow{2}{*}{\textbf{Method}} &
  \multicolumn{2}{l|}{\textbf{Detection}} & \multicolumn{2}{l|}{\textbf{Segmentation}} \\ \cline{4-7}
& & & \multicolumn{1}{c}{\textbf{AP}}       & \multicolumn{1}{c|}{\textbf{AP50}}       & \multicolumn{1}{c}{\textbf{AP}}         & \multicolumn{1}{c|}{\textbf{AP50}}        \\ \hline
\multirow{3}{*}{5} & \multirow{3}{*}{\checkmark} & One-Stage-Cosine & 5.37 & 9.91  & 4.3 & 7.32 \\ \cline{3-7}
&  & One-Stage-Linear     & 5.32 & 9.87  & 4.49 & 7.54 \\ \cline{3-7}
&  & iMTFA            & \multicolumn{1}{B{.}{.}{2,2}}{6.19} & \multicolumn{1}{B{.}{.}{2,2}|}{11.24} &\multicolumn{1}{B{.}{.}{2,2}}{5.22} & \multicolumn{1}{B{.}{.}{2,2}|}{8.71} \\ \hline
\end{tabular}}
\end{table}

\textbf{Effectiveness of the second fine-tuning stage.} To judge the merits of the second fine-tuning stage for feature extractor $\mathcal{G}$, we compare iMTFA to a variant that directly trains Mask R-CNN using a cosine-similarity head (One-Stage-Cosine) and one that directly uses the linear classification head in Mask R-CNN (One-Stage-Linear). iMTFA outperforms both, see Table~\ref{tab:extrafinetuning}. This demonstrates the effectiveness of the second fine-tuning stage. Instead of focusing on the cosine-similarity sub-network during training, One-Stage-Cosine seems to focus on the backbone and the RPN. One-Stage-Linear can produce embeddings with similar angles but dissimilar scales, which cannot be easily distinguished using cosine similarity.

\textbf{Cosine-similarity scaling factor.} \label{cosscaleeffect}
Parameter $\alpha$ in Eq.~\ref{alpha_equation} scales the classification scores before applying softmax. In Table~{\ref{tab:cosine-sim}}, we experiment with various $\alpha$ values and find that $\alpha = 1.0$ produces the best performance for COCO-Novel. For COCO-All, $\alpha = 10.0$ provides a good balance between high Overall and Novel AP. These values are subsequently used in all experiments on these datasets. For optimal performance, $\alpha$ needs to be tweaked based on the number of classes, which is in line with previous insights \cite{wojke2018deep}.

\begin{table}[H]
\centering
\footnotesize
\adjustbox{width=1.0\linewidth}{%
\begin{tabular}{|d{2.2}|c|c|c|c|c|c|}
\hline
\multicolumn{1}{|l|}{} & \multicolumn{2}{c|}{\textbf{COCO-Novel}}   & \multicolumn{4}{c|}{\textbf{COCO-All}}                                               \\ \cline{2-7} 
\multicolumn{1}{|l|}{} & \multicolumn{1}{c|}{\textbf{Detection}} & \multicolumn{1}{c|}{\textbf{Segmentation}} & \multicolumn{2}{c|}{\textbf{Detection}} & \multicolumn{2}{c|}{\textbf{Segmentation}} \\ \hline
\multicolumn{1}{|c|}{$\alpha$}               & \multicolumn{2}{c|}{\textbf{Overall}}      & \multicolumn{1}{c|}{\textbf{Overall}}    & \multicolumn{1}{c|}{\textbf{Novel}}    & \multicolumn{1}{c|}{\textbf{Overall}}      & \multicolumn{1}{c|}{\textbf{Novel}}     \\ \hline
1.0  & \textbf{6.22} & \textbf{5.24} & \multicolumn{1}{c|}{\textemdash}     & \multicolumn{1}{c|}{\textemdash}    & \multicolumn{1}{c|}{\textemdash}     & \multicolumn{1}{c|}{\textemdash}    \\ \hline
2.0  & 6.19     & 5.22     & 0.36  & 0.36 & 1.42  & 1.46 \\ \hline
3.0  & 6.17     & 5.21     & 10.31 & 6.17 & 9.55  & 5.21 \\ \hline
5.0  & 6.09     & 5.17     & 16.99 & \textbf{6.62} & 15.72 &\textbf{5.53} \\ \hline
10.0 & 5.76     & 4.94     & 19.62 & 6.07 & 18.22 & 5.19 \\ \hline
15.0 & 5.34     & 4.60      & \textbf{19.67} & 5.62 & \textbf{18.28} & 4.82 \\ \hline
20.0 & 4.94     & 4.26     & 19.45 & 5.19 & 18.09 & 4.47 \\ \hline
25.0 & 4.62     & 3.99     & 19.20  & 4.86 & 17.87 & 4.19 \\ \hline
\end{tabular}%
}
\caption{\textbf{Ablation alpha value.} Comparison of cosine scaling factors for iMTFA in COCO-All and COCO-Novel.}
\label{tab:cosine-sim}
\end{table}

\section{Conclusions} \label{sec:conclusion}

We have presented the first incremental approach to few-shot instance segmentation: iMTFA. iMTFA repurposes Mask R-CNN's feature extractor to generate discriminative per-instance embeddings. The mean of these embeddings is used as a class-representative in a cosine-similarity classifier. Because the localization and segmentation components are class-agnostic, the embeddings are all that is needed to add new classes. To compare iMTFA with a stronger non-incremental and class-specific baseline, we also introduced MTFA. It extends the few-shot object detection approach TFA \cite{wang2020frustratingly} by adding a mask prediction branch. Both iMTFA and MTFA outperform the current state-of-the-art on a variety evaluation scenarios using the COCO and VOC datasets.

There are several ways in which iMTFA can be improved. First, iMTFA cannot adapt to existing embeddings when generating new ones. Attention mechanisms such as those employed by \cite{gidaris2018dynamic,vinyals2016matching} are a promising future direction.

Second, iMTFA's class-agnostic localization and segmentation components are suboptimal compared MTFA's class-specific counterparts. An obvious improvement is to learn a transfer function from the generated embeddings to class-specific box regressor and mask predictor. We believe combining our approach with a guidance mechanism (e.g., \cite{fan2020fgn, michaelis2018one}) would further improve the performance of iMTFA.

Third, iMTFA's frozen box regressor and mask predictor introduce base class bias compared to MTFA. Employing guidance mechanisms would also alleviate this issue.

With these improvements in mind, the advances made with iMTFA present a promising outlook to narrow the gap between non-incremental and incremental few-shot instance segmentation, and to allow for a flexible addition of novel classes to already powerful networks.

\section{Acknowledgements} \label{sec:ack}

This work is supported by the Dutch Organization
for Scientific Research (NWO) with TOP-C2 grant ARBITER. We also thank Cyclomedia for sponsoring this research.

{\small
\bibliographystyle{ieee_fullname}
\bibliography{ms}
}

\end{document}